\documentclass[journal]{IEEEtran} 
\usepackage{cite}
\usepackage{amsfonts,amssymb} 
\usepackage{url}
\usepackage{amsmath}
\usepackage{amssymb}
\usepackage{graphicx} 
\usepackage{bm}
\usepackage{algorithmicx,algorithm}
\usepackage[noend]{algpseudocode}
\usepackage{booktabs}
\usepackage{multirow}

\ifCLASSINFOpdf
\else
\fi
\begin{document}
%
% paper title
% Titles are generally capitalized except for words such as a, an, and, as,
% at, but, by, for, in, nor, of, on, or, the, to and up, which are usually
% not capitalized unless they are the first or last word of the title.
% Linebreaks \\ can be used within to get better formatting as desired.
% Do not put math or special symbols in the title
\title{Multi-agent Soft Actor-Critic Based Hybrid Motion Planner for Mobile Robots}
%
%
% author names and IEEE memberships
% note positions of commas and nonbreaking spaces ( ~ ) LaTeX will not break
% a structure at a ~ so this keeps an author's name from being broken across
% two lines.
% use \thanks{} to gain access to the first footnote area
% a separate \thanks must be used for each paragraph as LaTeX2e's \thanks
% was not built to handle multiple paragraphs
%

\author{Zichen He, Lu Dong,~\IEEEmembership{~Member,~IEEE,} Chunwei Song, and  Changyin Sun,~\IEEEmembership{~Senior Member,~IEEE} % <-this % stops a space 

\thanks{Z.He and C.Song are with the College of Electronics and Information Engineering, Tongji University, Shanghai 201804, China (e-mail: 1910646@tongji.edu.cn; 2030739@tongji.edu.cn).}
\thanks{L.Dong is with the School of Cyber Science and Engineering, Southeast University, Nanjing 211189, China (e-mail:ldong90@seu.edu.cn)}
\thanks{C.Sun is with the School of Automation, Southeast University, Nanjing 210096, China, and also with the College of Electronics and Information Engineering, Tongji University, Shanghai 201804, China (e-mail: cysun@seu.edu.cn)}
}

\maketitle

% As a general rule, do not put math, special symbols or citations
% in the abstract or keywords.
\begin{abstract}
In this paper, a novel hybrid multi-robot motion planner that can be applied under non-communication and local observable conditions is presented. The planner is model-free and can realize the end-to-end mapping of multi-robot state and observation information to final smooth and continuous trajectories. The planner is a front-end and back-end separated architecture. The design of the front-end collaborative waypoints searching module is based on the multi-agent soft actor-critic algorithm under the centralized training with decentralized execution diagram. The design of the back-end trajectory optimization module is based on the minimal snap method with safety zone constraints. This module can output the final dynamic-feasible and executable trajectories. Finally, multi-group experimental results verify the effectiveness of the proposed motion planner. 
\end{abstract}

% Note that keywords are not normally used for peerreview papers.

% For peer review papers, you can put extra information on the cover
% page as needed:
% \ifCLASSOPTIONpeerreview
% \begin{center} \bfseries EDICS Category: 3-BBND \end{center}
% \fi
%
% For peerreview papers, this IEEEtran command inserts a page break and
% creates the second title. It will be ignored for other modes.
\IEEEpeerreviewmaketitle

\section{Introduction} 
%Model-free, CTDE, Hybrid Architecture, Front-end Path Searching, Back-end Trajectory Optimizing. Dynamic Feasible Trajectory at the Backend.   
%
%Local observable, non-communication. 
%Each agent cannot get the action info of others. 
%
%mSAC with automatic tuning alpha coeffs.   
% The very first letter is a 2 line initial drop letter followed
% by the rest of the first word in caps.
% 
% form to use if the first word consists of a single letter:
% \IEEEPARstart{A}{demo} file is ....
% 
% form to use if you need the single drop letter followed by
% normal text (unknown if ever used by the IEEE):
% \IEEEPARstart{A}{}demo file is ....
% 
% Some journals put the first two words in caps:
% \IEEEPARstart{T}{his demo} file is ....
% 
% Here we have the typical use of a "T" for an initial drop letter
% and "HIS" in caps to complete the first word.
\IEEEPARstart{I}{ntelligent} mobile robots are widely used to replace humans in performing complex, monotonous, and dangerous tasks due to their compactness, flexibility, and ability to carry a variety of sensors.  Nowadays, with the increasing complexity of operational tasks in the fields of warehousing, logistics, agriculture, subsea exploration, etc., there are apparent productivity and efficiency bottlenecks to operate with a single robot. For example, a single robot has limited perception capability and operating range. Multiple robots cooperating in the same space have advantages in efficiency, robustness, and task completion rate.  

Motion planning technology is critical to realize the autonomous collaborative operation of multiple robots.  The general description of multi-robot cooperative motion planning is: in the same space,  multiple robots need to start from different initial points to reach the corresponding target points.  The motion planner is required to plan a set of collision-free and dynamic-feasible trajectories with minimum time or energy consumption. Achieving such task with various interaction patterns is quite challenging, especially in dense environments where there exist multiple active agents under non-communication and local observable conditions. Most of the time, we have to consider dynamic constraints of different types of robots, including velocity, acceleration, curvature, jerk, snap, etc., so that trajectories are truly executable.  

\subsection{Related Works}
Multi-robot motion planning approaches can be divided into centralized methods and Decentralized methods.  Centralized methods contain optimization-based trajectory generation methods such as \cite{mellinger2012mixed,augugliaro2012generation}, and heuristic search-based planning methods such as \cite{yu2016optimal,svancara2017new}. Centralized methods have the advantage of completeness or probabilistic completeness.  However, this type of method requires obtaining global state information and cannot handle local observable situations.  In addition, centralized methods are constantly suffering from scalability issues.  As the number of robots in the task scenario increases, the computation complexity would rise exponentially, which is pretty challenging for the central computational device. 

\begin{figure}[!t]
	\centering
	\includegraphics[width=3.5 in]{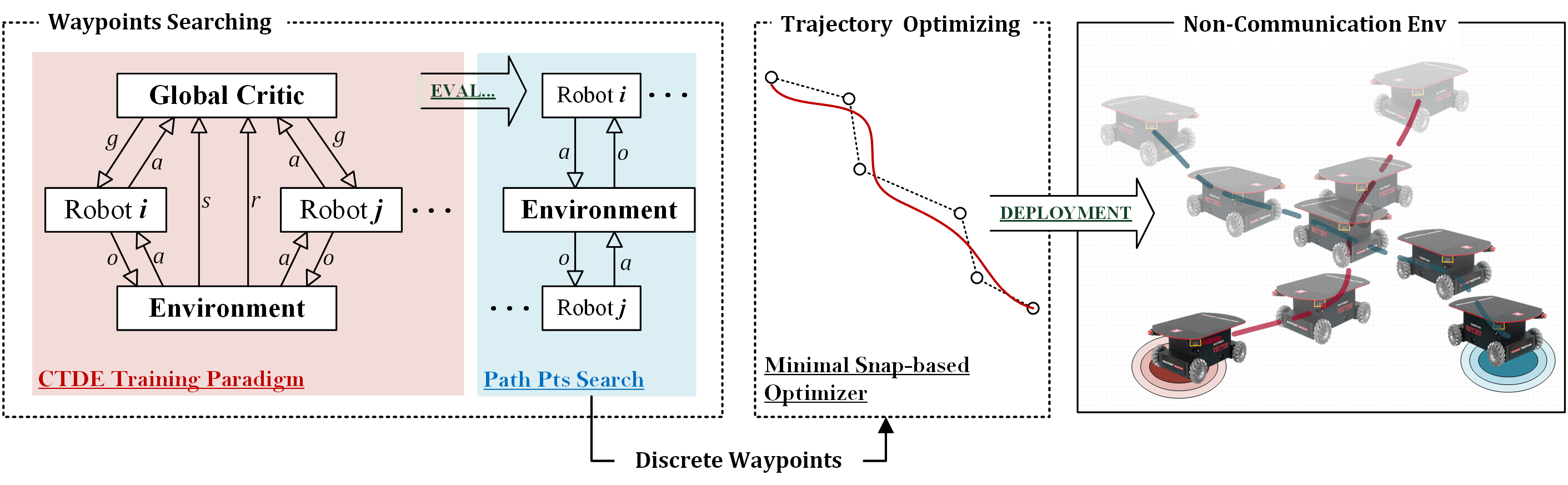}
	\caption{The architecture of the proposed MASAC-based hybrid motion planner. The CTDE RL paradigm-based front-end of this planner is responsible for searching the joint discrete waypoints. The minimal snap with constraints optimizer-based back-end of the planner is responsible for generating the final motion trajectory.}
	\label{Fig_TrajOptim} 
\end{figure}

In contrast, decentralized methods are widely studied for their stronger robustness and scalability.  The classical sampling path planning algorithm-based multi-robot motion planner is one of the mainstream research trends. In \cite{desaraju2011decentralized},   Desaraju et al. present DMA-RRT. This planner combines the rapid-exploring random tree (RRT) with the distributed multi-robot cooperative planning paradigm and can generate multiple paths for different robots.  However, DMA-RRT requires that each robot can communicate with others, and DMA-RRT does not consider the dynamic constraints of each robot.  In \cite{le2019multi}, Duong Le et al. utilize multi-agent search methods to guide the sampling-based planning process of each robot to effectively multi-robot motion planning problems with kino-dynamic constraints. The common problem of the above methods is that they do not separate the global and local planners. The whole planning process still relies on the priori map.

The velocity obstacle (VO) based methods are another type of decentralized multi-robot motion planning method that has been widely researched.  VO-based methods are based on reactive mechanisms and have the advantages of real-time and high efficiency.  VO does not consider the interaction pattern of other active agents in the same space \cite{douthwaite2018comparative}, which can cause the oscillation problem of final trajectories.  Reciprocal velocity obstacle (RVO) and its variants introduce implicit speed selection mechanisms and adopt probabilistic approaches to deal with uncertain interaction issues \cite{snape2011hybrid}. This operation can effectively solve the oscillation problem. Optimal reciprocal collision avoidance (ORCA) and its variants further improve the efficiency of real-time multi-robot trajectory generation.  In each time step, the robot obtains the velocity state of other robots and constructs its own speed space without collision.  The intersecting area from different speed spaces of robots constitutes the final optimal speed space.  Ultimately, the optimal joint motion strategy can be derived by solving linear programming problems \cite{van2011reciprocal}.  In \cite{douthwaite2018comparative},  scholars have demonstrated the superiority of ORCA over other VO-based methods in ideal scenarios.  ORCA can only handle motion planning problems where each robot is isomorphic and does not have non-holonomic constraints. In \cite{wang2018reciprocal,huang2019generalized},    researchers optimize the limitation of ORCA so that it can deal with more general task scenarios.  It is worth noting that there are many prerequisites for the application of ORCA and its variants. First, the robot has the perfect perception ability and can obtain the position, shape, and motion strategy information of $n$ robots within a specific range of itself. 

In recent years, with the development of machine learning,  learning-based model-free multi-robot motion planning methods have gradually become a research hotspot.  For example, in \cite{qureshi2020motion}, AH Qureshi et al. present motion planning networks (MPNet). MPNet is a neural motion planner based on the continuous learning paradigm and the expert supervision.  MPNet successfully constructs the mapping relationship between raw sensor streaming data and final bidirectional connected paths.  In \cite{riviere2020glas}. Riviere Benjamin et al. propose global-to-local safe autonomy synthesis (GLAS)  for multi-robot motion planning. GLAS integrates centralized methods with centralized methods. In GLAS, each robot acquires a distributed optimal policy through imitation learning. The expert experience of imitation learning comes from a priori centralized global optimal motion planner with resolution completeness.  The limitation of the above methods  is that the performance of the algorithm is depends too much on the quality of the priori expert demonstrations or the labeled dataset. 

Reinforcement learning (RL) approaches have received more attention than data-driven learning methods. RL is model-free and can learn in a ``trial-and-error" manner. Through interacting with the environments, the RL agent iteratively learns the policy to maximize the cumulative return \cite{he2021review,wang2021research}.  This pattern allows RL to integrate the learning process with the decision-making process and achieve end-to-end mapping from raw state input to the action output.   In the single mobile robot navigation task, RL-based motion planners have proven to deal with dense and dynamic scenarios without priori maps \cite{wang2018learning, shi2019end, he2021asynchronous}. Likewise, there are many scholars dedicated to researching multi-robot cooperative motion planning problems. Many studies have demonstrated that centralized training with decentralized execution (CTDE) -based multi-agent reinforcement learning (MARL) algorithms are pretty suitable for handling multi-robot collaborative planning problems \cite{chen2017decentralized, long2018towards}.  Compared to centralized MARL,  CTDE-based MARL has better scalability.  Besides, the centralized training process in CTDE avoids credit assignment and dynamic environment issues in decentralized MARL methods. In \cite{lowe2017multi, iqbal2019actor,yu2021surprising }, CTDE-based MARL algorithms such as MADDPG, MAAC, MAPPO have achieved great results in cooperative navigation scenarios in multi-agent particle simulation environments developed by OpenAI.  However, these works do not consider the trajectory shape and kino-dynamic constraints of real robots.    The ACL laboratory of MIT has made remarkable achievements in the field of  RL-based multi-robot motion planning \cite{chen2017decentralized, chen2017socially, semnani2020multi, everett2021collision}. In \cite{chen2017decentralized}, they propose collision avoidance with deep RL (CADRL) to solve multi-robot collaborative motion planning problems.  In \cite{chen2017socially}, they present socially aware CADRL (SA-CADRL) to address pedestrian-robot interaction issues on the basis of CADRL. Later, they consider the stochastic behavior model and introduce a supervised learning stage to solve the problem of dynamic environment information encoding \cite{everett2021collision}. In \cite{semnani2020multi} ,  they redesign the reward function and integrate original GA3C-CADRL with the force-based motion planning method to solve the long-range navigation problem.  The above researches promote the development of RL-based multi-robot motion planning. However, these works utilize the agent-level information of all robots. The observation input of each neural network contains policy information of other robots. 

To cope with the above limitations, we develop a hybrid motion planner suitable for non-communication and partial observable multi-robot motion planning tasks.  The specific architecture is shown in Fig. \ref{Fig_TrajOptim}. We combine the advantage of the CTDE-paradigm based MARL algorithm with the minimal snap-based trajectory optimization algorithm to establish an end-to-end mapping from local observations to the final executable, dynamic-feasible, smooth, and continuous trajectories of multiple robots.  In the front-end waypoints searching module,  we utilize the pre-trained MARL model to generate collaborative discrete waypoints. We extend soft actor-critic (SAC) to a MARL algorithm to handle the decentralized partially observable Markov decision problem (Dec-POMDP). In the back-end trajectory optimization module, we construct an optimization problem for solving the optimal coefficients of the continuous trajectory polynomials.  To sum up, our contributions are summarized as follows:

\begin{itemize}
	\item We propose a hybrid multi-robot motion planner that separates the front-end waypoints searching module and the back-end trajectories optimization module. This overall motion planner is model-free and does not rely on map priors.  
	\item We develop an off-policy multi-agent soft actor-critic-based cooperative waypoints searching method with self-adjusting policy entropy. This algorithm follows the CTDE paradigm. We set up multiple multi-robot motion planning scenarios under local observable and non-communication conditions and various baselines to verify the performance of this method. 
	\item We utilize a minimal snap trajectory optimization method with safety zone constraints to do the post-processing. This module performs shape adjustment and optimization for discrete paths and outputs smooth, continuous, and executable joint trajectories. Also, this separated module reduces the difficulty of offline MARL training and avoids the problem of non-convergence caused by introducing too many optimization goals in the RL framework.
\end{itemize} 

The remainder of this paper is organized as follows. The details of the front-end waypoints searching module and the back-end trajectory optimization module are described in Section II and Section III, respectively. Section IV presents the experimental results. Section V concludes this paper.

\section{Front-end Waypoints Searching}
In this section,  we describe in detail the front-end waypoints search module of the hybrid motion planner proposed in this paper. The specific content includes an introduction of the multi-agent soft actor-critic (MASAC) MARL framework based on the CTDE paradigm and a specification of the configuration of the state space, the action space, and the reward function in the multi-robot collaborative waypoints searching task.  

\subsection{Multi-agent Soft Actor Critic Framework} 
\subsubsection{Soft Actor Critic}
Before the soft actor-critic (SAC) algorithm is proposed, mainstream single agent model-free RL algorithms have some limitations. For example, the sample complexity of high-dimensional tasks leads to low sampling efficiency; a large number of hyper-parameters leads to unstable algorithm performance and weak generalization ability. The off-policy SAC balances sample utilization and algorithm stability. In addition,  the stochastic policy selection and the policy entropy mechanism are integrated into SAC. This operation enables SAC to encourage policy exploration by maximizing policy entropy, thus assigning nearly equal probability to those near-optimal actions with similar action-state values, avoiding repeatedly choosing the same action to fall into the sub-optimality. Meanwhile, the maximizing reward item ensures that the update process of the algorithm does not deviate from the overall optimization direction. Therefore, compared to DDPG, TD3, and other deterministic and continuous control RL algorithms, SAC has stronger policy exploration ability, generalization ability, and robustness, and thus is widely used in the field of robot learning.  

The core of SAC described above can be represented as follows:
% by equation (\ref{equation_sac}), 
\begin{equation}\label{equation_sac}
\pi_{\max }^{*}=\arg \max _{\pi} \sum_{t=0}^{T} E_{\left(s_{t}, a_{t}\right) \sim \rho_{z}}\left[r\left(s_{t},a_{t}\right)+\bm{\alpha} H\left(\pi\left(\cdot \mid s_{t}\right)\right)\right]
\end{equation}
where ${H(\pi(\cdot \mid s_t))=-\log \pi(\cdot \mid s_t)}$ is the policy entropy, which represents the degree of randomization of the policy. $\bm{\alpha}$ is the temperature coefficient, which represents whether the optimization objective of SAC is more inclined to maximize rewards or maximize policy entropy. ${\sum_{t=0}^{T} \mathbb{E}_{\left(\mathbf{s}_{t}, \mathbf{a}_{t}\right) \sim \rho_{\pi}}\left[r\left(\mathbf{s}_{t}, \mathbf{a}_{t}\right)\right]}$ is the basic maximization reward item in the objective function.  

\subsubsection{Multi-agent Soft Actor Critic Training Paradigm} 
The policy entropy maximization property and the hyperparameter insensitivity property of SAC determine that the agent naturally has certain robustness and generalization. This advantage makes SAC more suitable for robot learning with external disturbances and uncertain factors. Therefore, we propose a MARL training paradigm for the multi-robot waypoints searching method based on SAC.  

\begin{figure}[!t]
	\centering
	\includegraphics[width=3.5 in]{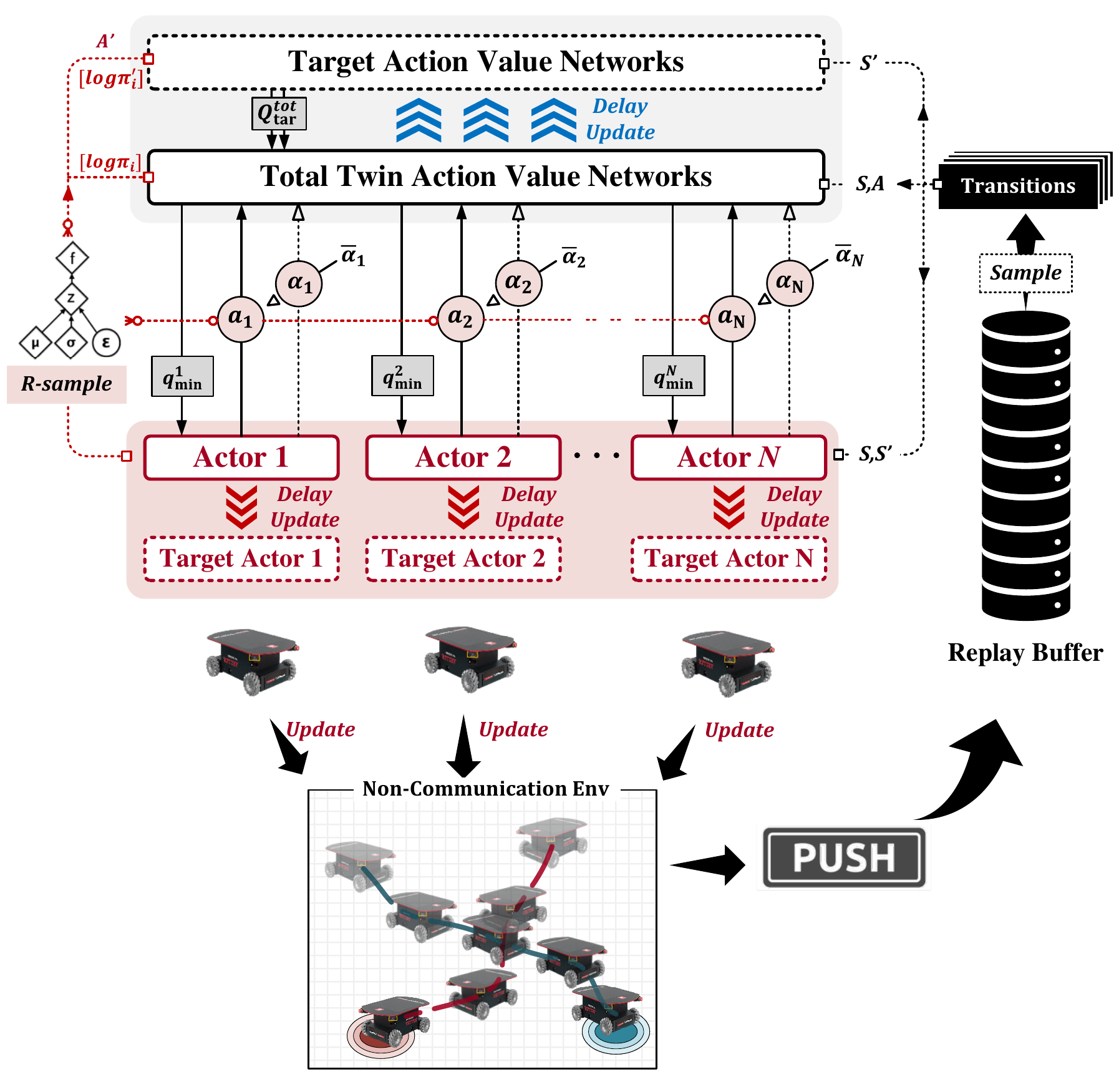}
	\caption{The detailed structure of MASAC off-policy training paradigm.}
	\label{Fig_TrajOptim} 
\end{figure}  

MASAC follows the CTDE paradigm. The robots represented by each actor are independent and cannot communicate with others. At each timestep, no robot can obtain the current motion policy of others. The advantage of CTDE is that there is no need to consider the credit assignment problem among agents. The optimization objective is to maximize the total reward, which is the sum of the rewards obtained by each robot after interacting with the environment. The reward function of each robot are independent of each other and only need to be set separately according to their own task requirements. The detailed structure of the MASAC training paradigm is illustrated in Fig. \ref{Fig_TrajOptim}. In this paradigm, the multi-robot waypoints search task can be modeled as a Dec-POMDP. 
\begin{equation}\label{DecPOMDP}
	G=\langle \bm{\hat{S}}, \bm{A}, \bm{P}, \bm{R}, \bm{\Omega}, \bm{I}\rangle
\end{equation} 
where ${i \in I=\{1,2, \ldots, N\}}$ represents the index set of each robot.  ${\bm{\hat{S}}=\langle \bm{S},\bm{O} \rangle}$ includes the global state and the collection of local observations of each robot. ${a_{i} \in A, \boldsymbol{a} \in A^{N}}$ represents the joint action of all robots at timestep ${t}$. ${\bm{R}=\{R_1,R_2,...R_N\}}$ is a tensor containing the reward signal of each agent. 
${P\left(s^{\prime} \mid s, \boldsymbol{a}\right)}$ is the transition function from the current state to the next state. ${o_{i} \in \bm{\Omega} \sim \bm{O}(s, i)}$ is the observation function. 

The global critic consists of two sets of action-value networks, one of which is a global twin soft Q network, and the other is the global target Q network that performs soft update based on the parameters of the twin soft Q network. Also, each actor includes a target policy network. The total twin soft Q Network outputs a pair of total Q values. We tend to take a smaller Q to weaken the overestimation bias in the Q learning process. The utilization of target networks of the global critic and actors is intended to stabilize the training process. 

The loss function of the global twin soft Q network is shown as follows:
\begin{equation}\label{Eq_LossFunQ} 
\begin{aligned}
	L(\theta_i) &=E_{\mathcal{D}}\left[\left(r_i\left(\boldsymbol{s}_{t}, \boldsymbol{o}_{t}, \boldsymbol{a}_{t}\right)+\gamma \min _{j \in 1,2} Q_{\theta_{j}}^{\prime t a r}\left(\boldsymbol{s}_{t+1}, \boldsymbol{o}_{t+1}, \tilde{\boldsymbol{a}}_{t+1}\right)\right.\right.\\
	&\left.\left.-Q_{\theta}\left(\boldsymbol{s}_{t}, \boldsymbol{o}_{t}, \boldsymbol{a}_{t}\right)\right)^{2}\right]
\end{aligned}
\end{equation}
where 
\begin{equation}\label{Eq_Qtar_prime}
\begin{aligned}
	Q_{\theta_{j}^{\prime}}^{\prime t a r}\left(\boldsymbol{s}_{t+1}, \boldsymbol{o}_{t+1}, \boldsymbol{a}_{t+1}\right) &=\mathrm{E}_{\pi^{i}}\left[Q_{\theta_{j}^{\prime}}^{\operatorname{tar}}\left(\boldsymbol{s}_{t+1}, \boldsymbol{o}_{t+1}, \tilde{\boldsymbol{a}}_{t+1}\right)\right. \\
	&\left.-\alpha_{t}^{i} \log \pi^{i}\left(\tilde{\boldsymbol{a}}_{t+1} \mid o_{t+1}^{i}\right)\right].
\end{aligned}
\end{equation} 
The joint trajectory ${\left(\bm{r}_t,\bm{s}_t, \bm{o}_t, \bm{a}_t, \bm{o}_{t+1}, \bm{s}_{t+1} \right)}$ is sampled from the replay buffer ${\mathcal{D}}$ at every specific timestep. ${\tilde{\bm{a}}_{t+1}}$ is obtained by re-sampling from current policy networks $\pi$. ${\gamma}$ is the discounted factor. ${\min _{j \in 1,2} Q_{\theta_{j}}^{\prime t a r}}$ is the twin soft Q values. We take the minimal one as the target. Equation (\ref{Eq_Qtar_prime}) illustrates its calculation process. In this equation, the policy entropy item is derived from each soft reward function 
\begin{equation}
r_i^{soft}\left(\bm{s}_{t}, \bm{o}_t,\bm{a}_{t}\right)=r_i\left(\bm{s}_{t}, \bm{o}_t,\bm{a}_{t}\right)+\gamma {\alpha}_t^i \mathbb{E}_{\pi^{i}} H\left(\pi_{tar}^i\left(\cdot \right)\right).
\end{equation}  
where ${Q_{\theta_{j}^{\prime}}^{\operatorname{tar}}}$ is the target Q network with parameters ${\theta^{\prime}}$. ${{\alpha}_t^i}$ is the exclusive temperature coefficient of each actor, which is used to control the weight of the policy entropy.  

The loss function of each policy network is shown as follows:
\begin{equation}\label{Eq_policyUpdate}
	\begin{aligned}
	L(\phi_i)&=\mathbb{E}_{\bm{o}_{t} \sim \mathcal{D}, \varepsilon \sim \mathcal{N}}\left[\alpha_t^i \log \pi_{\phi_i}\left(f_{\phi_i}\left(\varepsilon_{t} ; \bm{o}_{t}\right) \mid \bm{o}_{t}\right)\right. \\
	&\left.-Q_{\theta}\left(\bm{s}_t, \bm{o}_{t}, f_{\phi_i}\left(\varepsilon_{t} ; \bm{o}_{t}\right)\right)\right].
	\end{aligned}
\end{equation}
This function is derived from the soft policy iteration procedure. We also utilize the reparameterization trick here. For each sample of ${\pi_{\phi_i}(\cdot \mid \bm{o})}$, it is jointly determined by the current observation $\bm{o}$, parameters ${\phi}$, and the independent noise ${\varepsilon}$ that conforms to the standard normal distribution. We adopt ${f_{\phi_i}\left(\varepsilon_{t} ; \bm{o}_{t}\right)}$ to represent this squashing function. Its specific form is shown as follows:
\begin{equation}\label{Eq_SquashingGaussian}
	{f_{\phi_i}}(\bm{o}, \varepsilon )=\tanh \left(\mu_{\phi_i}(\bm{o})+\sigma_{\phi_i}(\bm{o}) \odot \varepsilon \right), \quad \varepsilon \sim \mathcal{N}(0, I)
\end{equation}
where ${\text{tanh}}$ activation function limits the final action to ${[-1,1]}$. $\mu_{\phi_i}$ and $\sigma_{\phi_i}$ are the outputs of the policy network. This trick allows us to optimize the policy parameters by computing the gradient of soft Q value directly. 

In addition, in order to better control the tradeoff between the exploitation and the exploration, and improve the parameters insensitivity property of the algorithm, we adopt an adaptive learning approach to control the temperature coefficient $\alpha_i$ of each actor. The purpose is to make each agent in MASAC pay more attention to exploring to improve the policy diversity in the early stage, and more inclined to utilize effective strategies to improve the training stability in the later stage. The objective function of $\alpha_i$ at timestep $t$ is as follows: 
\begin{equation}
	L(\alpha_i)=\mathbb{E}_{\bm{a}_{t} \sim \pi^i_{t}}\left[-\alpha_i \log \pi^i_{t}\left(\bm{a}_{t} \mid \tau_{t}\right)-\alpha_i \overline{\mathcal{H}}\right]
\end{equation}
$\overline{\mathcal{H}}$ is the target entropy,  usually set to the dimension of the action space ${-\text{dim}(\mathcal{A}_i)}$ of each actor. 

\subsection{RL Training Framework Configuration of Waypoints Searching Task}
The previous sub-section focuses on the description of the MARL framework based on the CTDE training paradigm. In this sub-section, we combine the functional requirements of the front-end waypoints search module in the hybrid motion planner to introduce our configuration methods for the action space, the observation space, and the reward function. 

The information contained in the continuous observation space of each mobile robot includes measurement data from onboard sensors and preset prior data. We hope that each mobile robot cannot directly obtain the action information of other robots, but learns to make inferences and predictions from limited local observation, and make its own optimal or near-optimal motion decisions. The specific configuration is as follows:
\begin{equation}
	\bm{o}_i = \left[i,\bm{v}_i,\bm{p}_i,\bm{\tilde{p}}_{\text{goal}},\bm{\tilde{p}}_{\text{others}},r_{\text{safe}}\right]
\end{equation}
where $i$ is the index number of current robot. $\bm{v}_i$ is the speed strategy of the robot $i$ at current timestep. $\bm{p}_i$ is the current global coordinate of the robot $i$ at current timestep. $\bm{\tilde{p}}_{\text{goal}}$ represents the relative coordinate of the goal with respect to the robot $i$. $\bm{\tilde{p}}_{\text{others}}$ represents the relative coordinate of the position of other robots with respect to the robot $i$. $r_{\text{safe}}$ denotes the radius of the rectangular-shaped safety zone of the robot $i$. Moreover, the global state configuration to input global soft Q network is as follows:
\begin{equation}
	\bm{s}_i = \left[ \bm{P}_{\text{robots}}, \bm{P}_{\text{goals}} \right]
\end{equation}
It contains global positions and corresponding target points of all robots at a specific timestep. 

The continuous action space of each mobile robots is set to $\bm{a}_i=\left[a_x, a_y\right] $. $a_x$ and $a_y$ are the components of the resultant acceleration generated by the robot actuator on the x-axis and y-axis, respectively. They are all in the value range of $[-1,1]$. In the actual deployment, the action range can change according to the complexity of the external environment.  

In this paper, we assume that each mobile robot is homogeneous and has the same task objective. We design the reward function of each mobile robot as follows: 
\begin{equation}\label{Eq_reward}
	r_t\left({\bm{o}_i,\bm{s}_i}\right)=\left\{\begin{array}{ll}
		1 & \text { if } d_g<0.1 \\
		-0.35 & \text { if } \text{any}(\bm{d}_{\text{robot}})<0 \\ 
		-0.35 + 0.05d_{robot}&\text{ if } 0<\text{any}(\bm{d}_{\text{robot}})<r_{th}\\ 
		d^{t-1}_g-d^{t}_g & \text { otherwise }.
	\end{array}\right.
\end{equation} 
where ${d_{g}=\left\|\bm{p}_{g}-\bm{p}\right\|_{2}}$ is the Euclidean distance between the current position of the robot $i$ and the target position. ${\bm{d}_{robot}}$ is the distance vector between the robot $i$ and other mobile robots. $r_{th}$ is the safety distance threshold between two mobile robots. It can be found that in addition to the final goal-reaching reward and collision penalty, we add many intermediate state reward signals to enable the robot to learn continuously and avoid some problems caused by the reward sparsity problem. 

Finally, after setting up MASAC training framework according to the above description and running the centralized training process with multiple episodes, the pre-trained actor model can be directly uploaded to the corresponding robot to perform decentralized collaborative waypoints online searching tasks without the global perfect sensing assumption. 

The details of training the MASAC-based front-end waypoints searching model are summarized in Algorithm \ref{Algortihm_Front_end}. 

\begin{algorithm}[htb]  
	\caption{MASAC Training Paradigm.}  
	\label{Algortihm_Front_end}  
	\begin{algorithmic}[1]  
		\State \textbf{Initialize} each policy network parameters ${[\phi_i]}_{i=1}^n$.  Initialize centralized twin soft Q networks   ${[\theta_i^1,\theta_i^2]}_{i=1}^n$. Initialize temperature coefficients of policy entropy ${[\alpha_i]}_{i=1}^n$. 
		\State \textbf{Initialize} replay buffers $\mathcal{D}$, independent noise $\varepsilon \sim \mathcal{N}(0,1)$. 
		\State \textbf{Initialize} parameters of target networks ${\phi}^{\prime}_i \leftarrow \phi_i, {\theta}^{\prime}_i \leftarrow \theta_i$.  
		\For {$eps = 1$ to $M$} 
			\If {$eps<M_{init}$}
				\State Warm-up stage.
			\EndIf
		\For{$t = 1$ to $eps_T$} 
			\State select action $\bm{a}^t$ according to equation (\ref{Eq_SquashingGaussian}). 
			\State $\bm{s}^{t+1} \sim p\left(\bm{s}^{t+1} \mid \bm{s}^{t}, \bm{a}^{t}\right)$. 
			\State add $( \bm{s}^t, \bm{o}^t, \bm{a}^t, \bm{r}^t, 
			\bm{o}^{t+1}, \bm{s}^{t+1} )$ to $\mathcal{D}$.  
			\If {$\text{update}$} 
				\State $\theta_i \leftarrow \theta_i - \alpha^{Q}_i \nabla_{\boldsymbol{\theta}_{i}} L(\theta_i) $ 
				\State $\phi_i \leftarrow \phi_i - \alpha^{\pi}_i \nabla_{\boldsymbol{\phi}_{i}} L(\phi_i) $ 
				\State $ \alpha_i \leftarrow \alpha_i - \nabla_{\boldsymbol{\alpha}_{i}} L(\alpha_i)$  
			\EndIf 
			\If {$\text{soft update}$} 
				\State $\theta_i^{\prime,j} \leftarrow \tau \theta_i^{j} - (1-\tau)\theta_i^{\prime,j}$, $ j \in [1,2] $ 
				\State $\phi_i^{\prime} \leftarrow \tau \phi_i - (1-\tau)\phi_i^{\prime}$ 
			\EndIf
		\EndFor 
			\State \textbf{end for}
		\EndFor 	
		\State \textbf{end for}
	\end{algorithmic}  
\end{algorithm}

\section{Back-end Trajectory Optimizing}

At present, in mainstream RL-based motion planning methods, mobile robots would directly track unsmooth folded segment trajectories formed by the discrete waypoints. However, in practical applications, the speed command and the acceleration of the mobile robot cannot change abruptly. This limitation makes mobile robots unable to follow the preset trajectory precisely, thus producing additional motion overshoot and resulting in safety risks such as collisions. Meanwhile, mobile robots need to accelerate and decelerate frequently  to track such polyline trajectories, which is extremely energy-consuming. If the trajectory generation and optimization are directly integrated into the RL architecture to realize the end-to-end mapping from state input to executable smooth trajectories, iterative reward function debugging processes have to be considered. Also, this method requires coupling kino-dynamic constraints and safety constraints of the robots, which undoubtedly increase the convergence difficulty of the algorithm. 

In this section, we separate the trajectory optimization process from the RL-based end-to-end motion planner, and utilize quadratic programming methods to generate optimal trajectories with minimal snap. This cascaded hybrid motion planning approach facilitates us to introduce kino-dynamics and safety constraints to solve smooth and executable trajectories with continuous velocity, acceleration, and jerk. Also, the operation of separating the trajectory optimization process avoids the introduction of a multi-objective reward form in the training phase, reduces the convergence difficulty, and allows the robot to focus on learning to obtain collision-free discrete waypoints.  Most importantly, from the front-end waypoints search to the back-end trajectory optimization, this hybrid motion planner helps us avoid the complex modeling process. 
% 这篇论文中，我们直接分离后端轨迹优化过程，基于前端生成的离散路径点，采用二次规划的方法来生成加速度变化最小的能量最优轨迹。这样既可以方便引入动力学等约束，解出速度、加速度连续的光滑可执行路径，同时可以降低训练端到端RL运动规划器方法中学习的复杂度。  

% Minimal Snap: 使得推力变化尽可能平缓，来最小化能量消耗。

\subsection{Trajectory Generation}
% 对任意两个路径点，可以用一段n阶多项式函数来描述它们的轨迹，即：
A piece of continuous trajectory formed by any two waypoints can be described by a segment of $n$-th order polynomial function: 
\begin{equation}\label{Eq_traj}
	f_i(t)=\left[1, t, t^{2}, \ldots, t^{n}\right] \cdot \bm{p}_i
\end{equation} 
where  
\begin{equation}
	\bm{p}_i=\left[p_{0,i}, p_{1,i}, \ldots, p_{n,i}\right]^{T}
\end{equation} 
${[p_{0,i},p_{1,i},...,p_{n,i}]}$ are the trajectory parameters, with the number of $n+1$. So, we can derive the speed, acceleration, jerk,  and snap representation of this two-point trajectory according to equation (\ref{Eq_traj}). The details are shown in equation (\ref{Eq_vajs}). 
\begin{equation}\label{Eq_vajs}
	\begin{array}{l}
v_i(t)=f^{\prime}(t)=\left[0,1,2 t, 3 t^{2}, 4 t^{3}, \ldots, n t^{n-1}\right] \cdot \bm{p}_i \\
a_i(t)=f^{\prime \prime}(t)=\left[0,0,2,6 t, 12 t^{2}, \ldots, n(n-1) t^{n-2}\right] \cdot \bm{p}_i \\
\text{jerk}_i(t)=f^{(3)}(t)=\left[0,0,0,6,24 t, \ldots, \frac{n !}{(n-3 !)} t^{n-3}\right] \cdot \bm{p}_i \\
\text{snap}_i(t)=f^{(4)}(t)=\left[0,0,0,0,24, \ldots, \frac{n !}{(n-4 !)} t^{n-4}\right] \cdot \bm{p}_i
\end{array}
\end{equation} 
Now, we assume there exists $M$ mobile robots. For any robot $m$, there are $K+1$ waypoints generated by the MARL-based front-end. Therefore, the whole motion trajectory of this robot composed of $K$-segment polynomials for this robot has the following representation: 
\begin{equation}
	\begin{aligned}
	f(t)=\left\{\begin{array}{ll}
f_{1}(t) \doteq \sum_{i=0}^{n} \bm{p}_{1, i} t^{i}   &T_{0} \leq t \leq T_{1} \\
f_{2}(t)  \doteq \sum_{i=0}^{n} \bm{p}_{2, i} t^{i}  &T_{1} \leq t \leq T_{2} \\
\vdots  \\
f_{K}(t)  \doteq \sum_{i=0}^{n} \bm{p}_{K, i} t^{i}  &T_{K-1} \leq t \leq T_{K}
\end{array}\right.
	\end{aligned}
\end{equation} 
where $T$ is the time node of each segment. 

\subsection{Minimal Snap with Safety Zone Constraints} 
% 首先，结合上面分析，我们的优化目标是，在诸多约束下，选择一组最优的参数组合P=[PK1,1,PK2,2,PK3,3] 使得多个移动机器人协同运动规划的运动轨迹的Snap最小，即机器人执行器推力变化尽可能平缓，从而最小化能力消耗。
Combined with the above derivation, our optimization objective is to select a set of optimal polynomial parameter combinations $P=[P_{R1},P_{R2},...,P_{RM}]$ (where $P_{Rm}=[\bm{p}_1^m,\bm{p}_2^m,...,\bm{p}_{K}^m]$) under various constraints to minimize the snap of the trajectory of each robot. In this way, the actuator thrust of each robot changes as smoothly as possible, thereby minimizing energy consumption. 

For a single trajectory of the mobile robot $m$, the cost function of minimal snap can be written as:
\begin{equation}
\begin{aligned}
J(T) &=\min \int_{0}^{T}\left(f^{(4)}(t)\right)^{2} \mathrm{d} t=\min \sum_{i=1}^{K} \int_{T_{i-1}}^{T_{i}}\left(f^{(4)}(t)\right)^{2} \mathrm{d} t \\
&=\min \sum_{i=1}^{K} \bm{p}_i^{T} \bm{Q}_{i} \bm{p}_i
\end{aligned}
\end{equation} 
where $f^{4}(t)$ can be obtained in equation (\ref{Eq_vajs}). $\bm{p}_i$ represents the polynomial parameters of each segment. $\bm{Q}_i$ is the Hessian matrix, the detail is as follows:
\begin{equation}
\bm{Q}_i=\left[\begin{array}{cc}
0_{4 \times 4} & 0_{4 \times(n-3)} \\
0_{(n-3) \times 4} & \begin{array}{cc}\frac{r(r-1)(r-2)(r-3)c(c-1)(c-2)(c-3)}{r+c-7}\times\\ \left(t_{i}^{(r+c-7)}-t_{i-1}^{(r+c-7)}\right)
\end{array}\end{array}\right]_{r\times c}
\end{equation}
where $r$ and $c$ respectively represent the number of rows and columns of $Q_i$. 

%在本文所描述的任务场景中，每个移动机器人的轨迹优化过程的约束包括等式约束与不等式约束。 
In the task scenario described in this paper, the constraints of the trajectory optimization process of each mobile robot include several equality constraints and several inequality constraints. 
\subsubsection{Equality Constraints}
% 首先考虑到初始移动机器人的位置、速度、加速度，jerk的状态约束：
First, we introduce the initial and terminal state constraints of the mobile robot, including  position, speed, and acceleration: 
\begin{equation}
f^{(d)}\left(T_{0,T}\right)=s_{0,T}^{(d)}.
\end{equation}
We transform it into the standard input form of the QP optimizer:
\begin{equation}
	\bm{A}_{0,T}\bm{p}_{0,T} = \bm{s}_{0,T}
\end{equation} 
where $\bm{A}_{0,T} = [\bm{A}_0,\bm{A}_T]^T$ is the mapping matrix between polynomial parameters and state, $\bm{s}=[x,v,a]^T$.  

% 另一类等式约束为连续性约束，即条件状态约束来保证轨迹中间相邻片段之间的连续性：
The other type is the continuity constraint. We ensure the continuity of the whole trajectory by constraining endpoint derivatives of the segment $i$ to be equal to initial derivatives of the segment $i+1$: 
\begin{equation}
\left.\begin{array}{l}
f_{i}^{(d)}\left(T_{j}\right)=f_{i+1}^{(d)}\left(T_{j}\right) \\
\Rightarrow\left[\bm{A}^{i}-\bm{A}^{i+1}\right]\left[\begin{array}{c}
\bm{p}_{i} \\
\bm{p}_{i+1}
\end{array}\right]=0
\end{array}\right.
\end{equation} 

\subsubsection{Inequality Constraints} 
% 首先需要考虑选用机器人执行器的限制，需要使得轨迹全程速度大小与加速度大小在允许的范围内，即设置
First, we need to set ${v_{min}, v_{max}, a_{min}, a_{max}}$ to constrain the motion speed and acceleration of each mobile robot. Moreover, we introduce a set of rectangular-shaped safety zones to constrain the middle points of each segment in the trajectory instead of fixing each middle point according to the classical minimal snap method. In detail, we perform multiple middle path point sampling processes in every intermediate segment. For each path point, we add two range inequality constraints on the x-axis and y-axis: 
\begin{equation}
	\begin{array}{c}
{A_{T_j} \bm{p}_i \leq f_i\left(T_{j}\right)+d_{\text{safe}}} \\
{-A_{T_j} \bm{p}_i \leq-\left(f_i\left({T_j}\right)-d_{\text{safe}}\right)}
\end{array}
\end{equation}
This corridor-based soft constraints approach contains an implicit time allocation mechanism. Furthermore, it avoids the problem of overshoot when optimizing trajectory by the classical minimal snap method with solid constraints. 

% 另外，我们额外设置了rectangular shaped safety zone来约束轨迹的中间点取代了传统方法中对中间点进行固定。具体做法是，我们在轨迹中间每一段中进行多次采样，并且对于每一个点，我们添加一个安全区域的不等式约束，来避免经典强约束中间点的轨迹优化过程中的超调问题。 

\begin{table}[!t]
	\centering
	\caption{Hyper-parameters configuration of MASAC training process}
	\setlength{\tabcolsep}{5mm}
	\label{Table_MASACTraining}
	\begin{tabular}{lc}
		\toprule[1pt]
		\textbf{Parameters}   & \textbf{Value}                     \\ \hline
		$\bm{\alpha}$(init)        & [0.01,0.01,0.01]                               \\
		target entropy        & [-dim(12),-dim(12),-dim(12)]                          \\
		learning rate(actor)  & 0.001                               \\
		learning rate(critic) & 0.001                              \\
		optimizer             & Adam                               \\
		tau                   & 0.005                              \\
		mini-batch size       & 2048                               \\
		episodes              & 5e4                                \\
		warm-up episodes      & 1e3                                \\ 
		memory length         & 1e6                                \\ 
		timesteps per episode & 50								  \\ 
		update frequency      & 50                                \\
		actor delay frequency & 1                                  \\ 
		$d_{\text{safe}}$(m)         & 0.3 							   \\ 
		\bottomrule[1pt]
	\end{tabular}
\end{table}

\begin{figure}[!t]
	\centering
	\includegraphics[width=3.0 in]{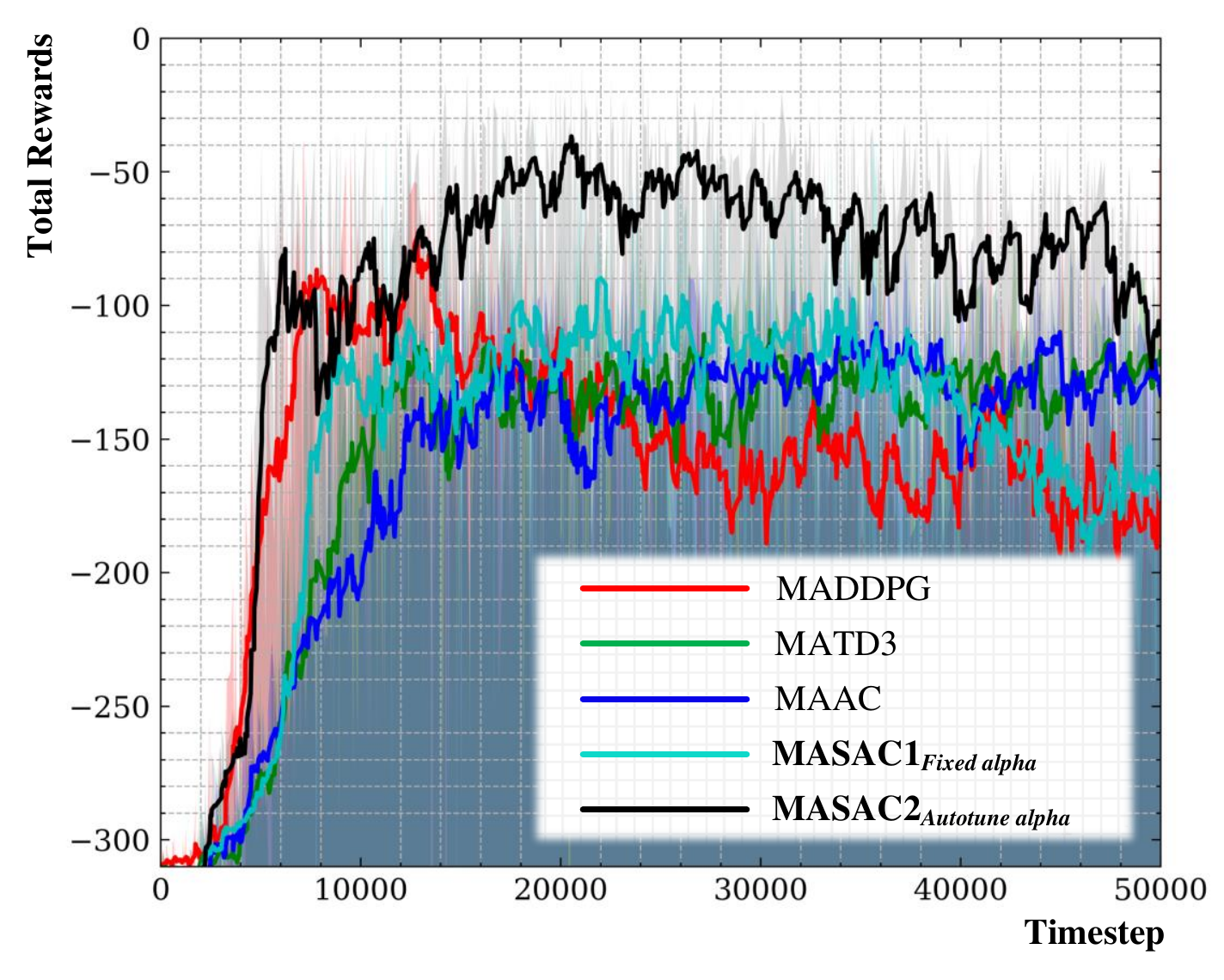}
	\caption{The average reward curves for MASAC, MATD3, MADDPG, and MAAC on the 3-robot waypoints searching task scenario.}
	\label{Fig_RewardCurves} 
\end{figure}   

\begin{figure*}[!t]
	\centering
	\includegraphics[width=5.0 in]{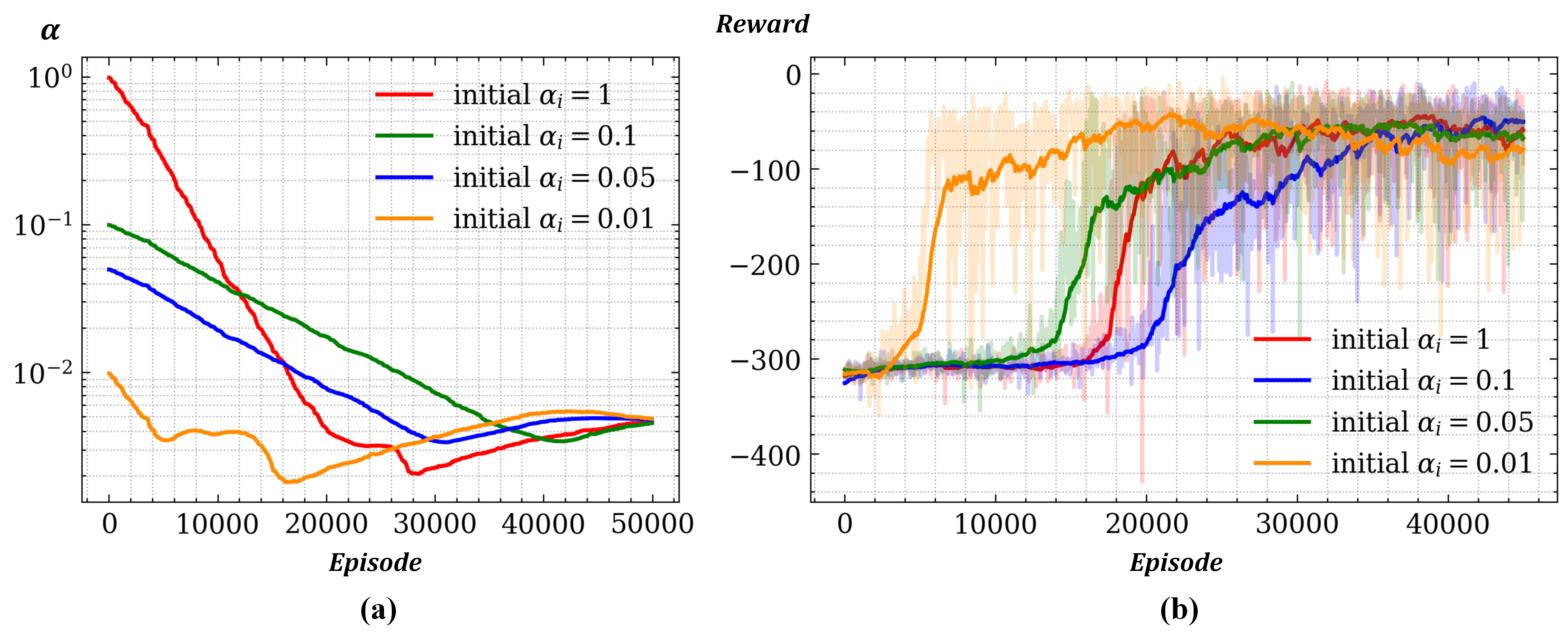} 
	\caption{(a) represents the self-learning process of average temperature coefficients of different agents with different initial values. (b) represents the average reward curves of MASAC methods with different initial $\alpha$ during the training process. }
	\label{Alpha_tuning}
\end{figure*}   

%Combining the descriptions of the front-end waypoints searching module and the backend-end trajectory optimization, the details of the hybrid motion planner proposed in this paper are summarized in Algorithm \ref{alg2_hybrid_motionplanner}  
%\begin{algorithm}
%	%\textsl{}\setstretch{1.8}
%	\renewcommand{\algorithmicrequire}{\textbf{Input:}}
%	\renewcommand{\algorithmicensure}{\textbf{Output:}}
%	\caption{MASAC-based Hybrid Motion Planner}
%	\label{alg2_hybrid_motionplanner}
%	\begin{algorithmic}[1]
%		\State 
%	\end{algorithmic}  
%\end{algorithm}

\begin{figure*}[]
	\centering
	\includegraphics[width=6 in]{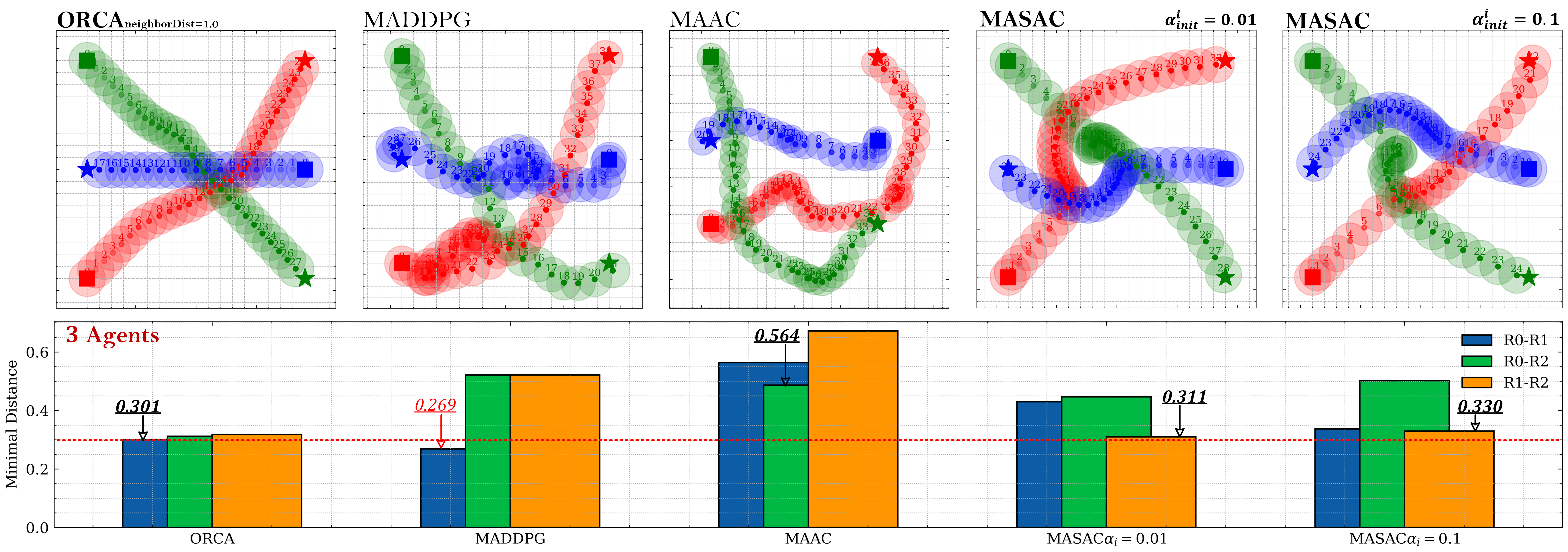}
	\caption{The visualization inference results of different algorithms in the unknown 3-robot fully interaction planning scenario. The rightmost bar chart compares the minimal distance between robots during the planning process.}
	\label{Fig_Comparision3agents} 
\end{figure*}  

\begin{table*}[!t]
	\centering
	\caption{Performance metrics comparison of 3 robots collaborative path planning based on different algorithms}
	\label{Tab_all_metrics}
	\begin{tabular}{cccccc}
		\toprule[1pt]
		\textbf{Algorithms}                         & \multicolumn{5}{c}{\textbf{Different Performance Indicators}}                                                                         \\ \hline
		\multicolumn{1}{l}{\textbf{Motion Planner}} & \textbf{Avg Total Distance} & \textbf{Avg Total Time} & \textbf{Avg Total Reward} & \textbf{Collsion Numbers} &\textbf{Success Rate} \\ \hline
		ORCA                                        & 1.928                       & 114.538                 & -                         & 10(9049)                  & 73.627\%              \\
		MATD3                                      & 3.497                       & 124.013                 & -28.804                   & 25(9673)                  & 92.301\%              \\
		MAAC(CAS)                                   & 3.209                       & 123.679                 & -33.695                   & 14(9647)                  & 92.051\%              \\ \hline
		\textbf{MASAC-auto}                         & \textbf{2.969}              & \textbf{118.897}        & \textbf{-39.545}          & \textbf{2(9274)}          & \textbf{93.592\%}     \\ \bottomrule[1pt]
	\end{tabular}
\end{table*}

\begin{figure*}[!t]
	\centering
	\includegraphics[width=5.5 in]{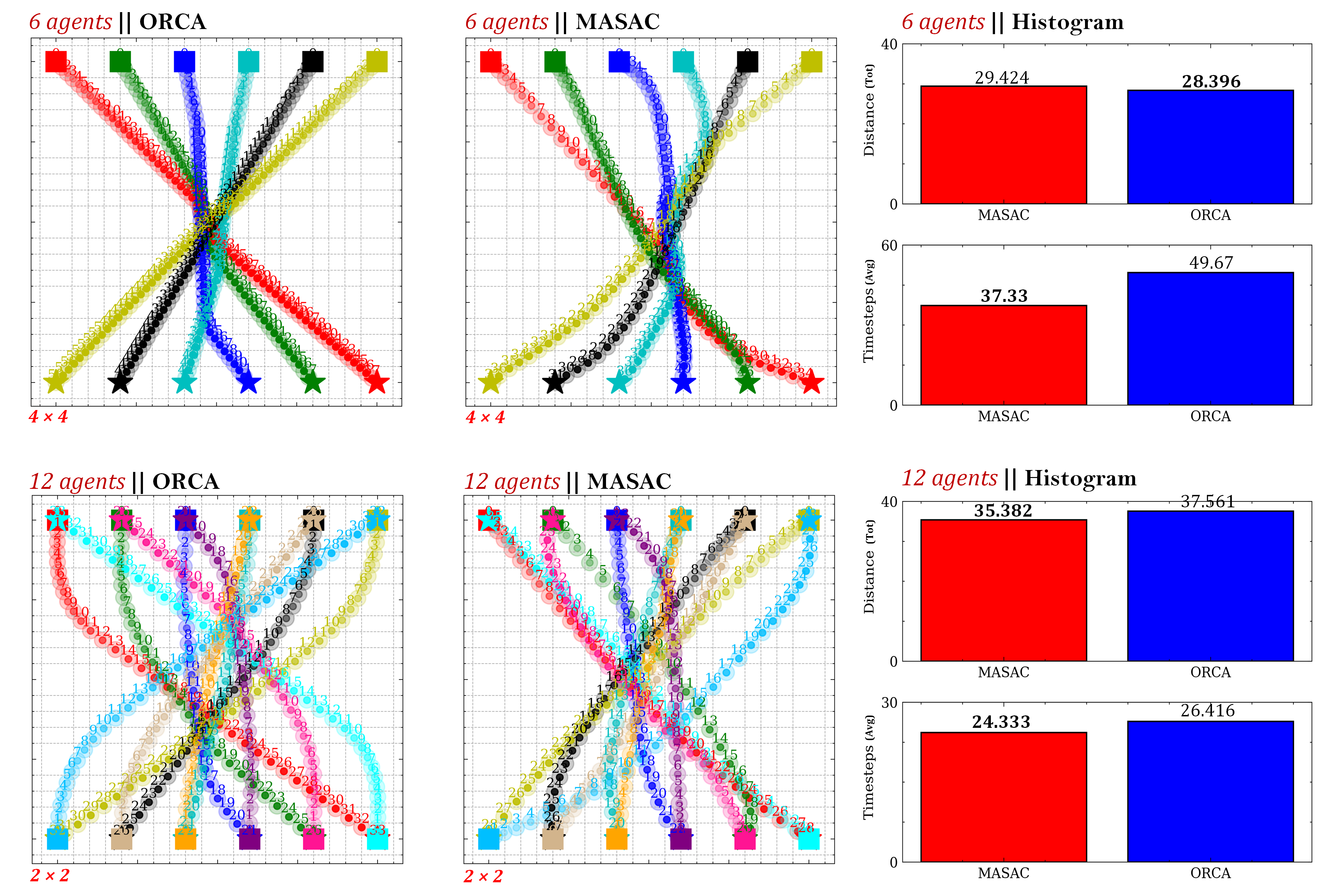}
	\caption{6-robot and 12-robot planning scenarios. ORCA is still selected as an ideal comparison algorithm. The total planning distance length and average time consumption metrics of these two methods are compared in the rightmost bar charts.}
	\label{Fig_6agentsAnd12agents} 
\end{figure*}

\section{Experiments}
In this section, we first present the training process and the collaborative planning effect of the MASAC-based front-end waypoints searching module, and analyze the comparison results between this algorithm with multiple model-free algorithms with the CTDE paradigm and a mainstream model-based algorithm. Then, we combine the pre-trained front-end waypoints module with the back-end trajectory optimization module, and demonstrate the trajectory planning effect of the final hybrid motion planner and make some analysis.

% 这一部分中，我们首先展示了前端路径点搜索模块的训练过程与训练与泛化效果，并设置了多组的无模型算法与一组主流的基于模型的算法进行对比。接着我们将前端路径路径点搜索模块与后端轨迹优化模块结合，并展示了最终混合规划器的规划效果，并进行了一些分析。 
\subsection{Experiments of the Front-end Waypoints Searching Module}
% 我们的算法的基准测试场景是多个给定目标点的移动机器人的协同规划任务。在对比组算法的选取上，我们选择了同样是集中训练分布执行架构下基于MADDPG算法的改进算法MATD3，以及基于注意力机制的MAAC算法。并且，我们额外设置了一组主流的基于模型的多机器人协同路径规划算法作为对照。值得注意的是，ORCA是一种理想算法，该算法的前置条件是所有机器人具有相同的构型，且机器人之间是完美通信的，即每个机器人都知道其他机器人当前时刻的策略。而我们所提出的方法中，每个机器人是部分可观的，机器人之间不存在通信，无法获取其他机器人的当前动作策略。这组对照方便我们对比本算法的规划效果与可通信算法之间的性能差异。
We select the multi-robot collaborative navigation task with preset target points as the benchmark test scenario. As for the selection of the comparison baselines, we adopt state-of-the-art MATD3 (an improved algorithm on the basis of MADDPG) and MAAC algorithms which are all training under the CTDE paradigm. Furthermore, we set optimal reciprocal collision avoidance (ORCA) as the ideal baseline. ORCA is a commonly used multi-robot interaction algorithm based on the velocity obstacle (VO). ORCA is an ideal algorithm. It has several preconditions. First, all robots should be homogeneous. Besides, there exists a perfects communication assumption between robots, i.e., each robot can obtain the motion strategy of other robots at each timestep. This baseline facilitates us to compare the performance difference between our model-free and local observable method with the communicable method.

%我们路径点搜索模块的MASAC架构的中心Critic包括两个Q网络，每个Q网络包含3个MLP层，每个MLP层的神经元数量为[1024,512,300]。策略网络包括2个MLP层，每个MLP层的神经元数量为[500,128]。训练的Trick，我们为训练过程配置了warm-up的训练阶段，以及Reward scaling过程来稳定训练，同时采用了自学习的温度系数来平衡探索与利用。具体的超参数配置见表1.  
The centralized critic of the MASAC architecture of our waypoints searching module consists of two soft Q networks. Each soft Q network contains three fully connected hidden layers, and the number of units in each layer is $[1024,512,300]$. Each distributed policy network contains two fully connected hidden layers, and the number of units in each layer is $[500,128]$. During the training process, we introduce the warm-up training stage and reward scaling trick for stability purposes. Also, we assign an independent temperature coefficient to each actor and utilize the self-tuning approach to balance the exploration and exploitation. At the beginning of each training phase, we randomly initialize the positions of every robot and corresponding target point. The specific hyperparameters are configured as shown in Table \ref{Table_MASACTraining} (3-robot waypoints searching task). 

% CPU信息 Intel(R) Xeon(R) Silver CPU 
% GPU信息 NVIDIA RTX3090 
% 
We deploy all algorithms on a computer with AMD Ryzen7 5800H CPU and NVIDIA RTX 3060 GPU for training. We take the 3-robot waypoints search task as an example and list the average reward curves of different algorithms. The details are visualized in Fig. \ref{Fig_RewardCurves}. It can be found that MASAC with multiple self-learning temperature coefficients $\bm{\alpha}=[\alpha_1,\alpha_2,...,\alpha_n]$ can speed up the training process and help the robot obtain the maximum average reward without communication. Also, during the hyper-parameter-tuning process, we find that the initial value of the exploration temperature coefficient $\alpha$ has a certain impact on the final performance of MASAC. As shown in Fig. \ref{Alpha_tuning}, the subplot (a) shows that MASAC can indeed effectively adjust the exploration strength during the training process. Also, different initial $\alpha$ eventually converges to similar final states. Subplot (b) illustrates that when the initial value of each $\alpha_i$ is 0.01, MASAC has a faster convergence speed compared to the other three groups. However, it should not be ignored that a slight performance degradation occurs at the end of the training phase. A comprehensive comparison reveals that the degree of this decay disappears with increasing of the initial $\alpha_i$ of each agent.  

% 并且，在研究过程中，我们发现alpha的初值会对算法表现产生较大的影响。因此，我们对比了不同alpha时算法的训练表现。如图figure. \ref{} 所示。从[a]子图中可以发现，在训练过程中，算法确实可以有效的进行探索力度的调整。并且，不同初值最终也会收敛到类似的终态。从[b]子图中可以发现，当alpha=0.01时，相较于另外三组，算法整体有着更快的训练速度，但是最训练默契，会出现轻微的性能退化。

% 推理测试如Fig所示，我们设计了一种能使3个移动机器人充分交互的协同路径点搜索的场景作为测试基准。 可以看出，在已知其他机器人运动策略的理想条件下，ORCA可以以接近机器人之间最小安全阈值的形式进行交互，从柱状图也可以看出，在交互期间，机器人之间的最小距离为0.301。而在无通信的部分可观场景下，MATD3架构下机器人在交互期间的最小距离为0.269，在第11个时间步，红色机器人与绿色机器人发生了碰撞。收敛到次优策略的MAAC算法相对比较保守，在交互期间及机器人的最小间距为0.564。MASAC相对取得比较符合直觉的交互结果，机器人最小相对间距为0.311. 从图中也可直观看出比较接近理想ORCA的交互方式。 
The real-time inference test results are summarized in Fig. \ref{Fig_Comparision3agents}. We design an unknown scenario that enables each mobile robot could fully interact with others as an inference test benchmark. The results show that ORCA allows robots to interact with an approximate minimum safety threshold (0.301) under the condition of perfect communication. On the other hand, MATD3, MAAC, and MASAC are running under non-communication and partial observation conditions. The minimum distance between robots during the interaction process of MATD3 is 0.269. It can also be verified in the second sub-figure that the red robot collides with the green one at the 11-th timestep. The robots trained in MAAC architecture are relatively conservative. They tend to select suboptimal motion strategies to ensure a sufficient safety distance (0.564). 

The other two MASAC-based waypoints searching methods obtain more intuitive collaborative planning results. Under the premise that other robot motion strategies cannot be obtained, each robot chooses to slow down or bypass in advance to avoid others. When we set the initial exploration temperature coefficient $\alpha_i = 0.01$, the robot behaves more conservatively (0.311). In contrast, the final result is closer to the ORCA when we adjust $\alpha_i$ to $0.1$ (0.330).  This is because a bigger exploration degree in the early stage of the training phase will increase the complexity and diversity of samples, which will sacrifice part of the training speed. However, the algorithm will therefore have better robust performance and inference generalization ability.

% 进一步地，我们统计了100次随机生成场景的协同路径点搜索的结果，并设置了多组评价指标，包括规划的平均距离，平均时间耗费，平均奖励回报，100次所有机器人耗费总时间步数以内的碰撞次数，以及在有限的50个timestep内的规划成功率（所有机器人都达到目标点就记为成功）。所有结果汇总成下表2。可以发现，MASAC的综合性能最接近理想的ORCA算法，并且在有限的时间步长的情况下，规划的成功率最高。 

Further, we design an algorithm to randomly generate 100 task scenarios of 3-robot waypoints searching for evaluation. Also, we selected a variety of evaluation metrics, including the average distance sum of robots, the average time consumption of robots,  the average reward sum,  total collision numbers, and the search success rate within limited timesteps (Note that ``success" is recorded when all robots reach their target points.) All results are aggregated in Table \ref{Tab_all_metrics}. This result suggests that MASAC has better comprehensive waypoints searching performance and is closer to the performance of the ideal ORCA algorithm compared to MAAC(continuous action space version) and MATD3. Moreover, the MASAC-based searching method has the highest success rate with limited timesteps.

% 泛化性能测试上，我们又额外设置了更复杂的交互场景，具体如下图5所示。第一种是6个机器人单向交互的路径点搜索场景，第二种是12个机器人双边双向的交互的路径点搜索场景。我们同样以理想的ORCA算法作为对比基准，可以发现，MASAC算法在该场景的规划距离长度上已经接近ORCA的水准，并且消耗的规划时间步更少。 
For further evaluating the inference ability of the proposed method, we set up more complex unknown fully interaction scenarios. The details are shown in Fig. \ref{Fig_6agentsAnd12agents}. The first waypoints searching scenario contains six robots. All robots interact fully from the same side to the other side. The second one is a 12-robot bilateral bi-directional interaction waypoints searching scenario.  It should be noted that in the 12-robot environments, we choose a smaller scenario scale to increase the difficulty of inference. We also utilize optimal ORCA(Global distance perception) as a comparison benchmark. The aggregated results in this figure suggest that the MASAC-based method has great scene generalization ability. In the 6-robot long-distance planning scenario, the proposed method approaches the performance of perfect sensing ORCA in the total distance length indicator, and has less timestep consumption. In the narrower 12-robot scenario, ORCA with the real-time policy sharing mechanism makes each robot go around a distance in advance. In contrast, our method achieves better results on the matrics of total distance length and average timestep consumption. 

\subsection{Experiments of the Hybrid Motion Planner} 

\begin{table}[!t]\label{Table_HybridMotionPlanner}
	\centering 
	\setlength{\tabcolsep}{5mm}
	\caption{Hyper-parameters configuration of minimal snap trajectory optimization with safety zone constraints}
	\label{Table_TrajOptimProcess}
	\begin{tabular}{lc}
		\toprule[1pt]
		\textbf{Parameters}            & \multicolumn{1}{l}{\textbf{Value}} \\ \hline
		order $n$                         & 5                                  \\
		length of the safety zone $l$     & 0.1                                \\
		width of the safety zone $w$      & 0.1                                \\
		intermediate sampling interval $r$ & 0.05                               \\
		max iteration  $N$               & 1000                               \\ 
		action range $[a_{min},a_{max}]$   & 
		[-1.0,1.0]                         \\ 
		\bottomrule[1pt]
	\end{tabular} 
\end{table}

% 在以基于MASAC的前段路径点搜索模块的基础上，我们加入了基于Minimal Snap的轨迹优化方法，来生成机器人速度、加速度光滑且连续的可执行能量最优轨迹。并且我们引入安全区域的不等式约束来代替中间点的Fix强约束，这引入了隐形的时间分配机制，并有利于轨迹形状的修正。详细做法如Section III所述。具体后端轨迹优化过程的参数配置汇总见下表：
Based on the front-end MASAC waypoints searching module, we integrate the minimal snap trajectory optimization method with safety zone constraints. The final hybrid motion planner can generate dynamic-feasible, collision-free, and energy-optimal trajectories for multiple mobile robots navigating cooperative motion planning under no-communication and partial observation conditions. Besides, the velocity and the acceleration profiles of generated trajectories are smooth and continuous. This feature is conducive to the design of the low-level tracking controllers.  The hyperparameters configuration of the back-end trajectory optimizing module is summarized in Table \ref{Table_Backend}.  

% 以三个智能体的充分交互场景为例。我们预设每个机器人的初始速度、加速度，以及到达目标点后的终止速度，加速度均为0。同时我们选择经典的Minimal Snap方法作为对比基准。最终生成的轨迹图如下图Fig所示。可以看出，在无通信且部分可观的前提下，相较于基于经典Minimal Snap方法的运动规划器，带有安全区域约束轨迹优化模块的混合运动规划器生成的最终轨迹更加的光滑。归功于隐式的时间分配机制，该混合规划器在保证满足在安全区域约束的前提下，采用了更加合理的速度管理方法，修复了无效的轨迹形状。
We take the 3-robot fully interactive scenario as an example. We preset the initial speed, initial acceleration, terminal speed, and terminal acceleration to the zero states. Meanwhile, we select the classical minimal snap method as the comparison benchmark. We aggregate the final trajectory generation results into the following Fig. \ref{Fig_FinalTraj_Comparsion}.
\begin{figure}[!t]
	\centering
	\includegraphics[width=3.5 in]{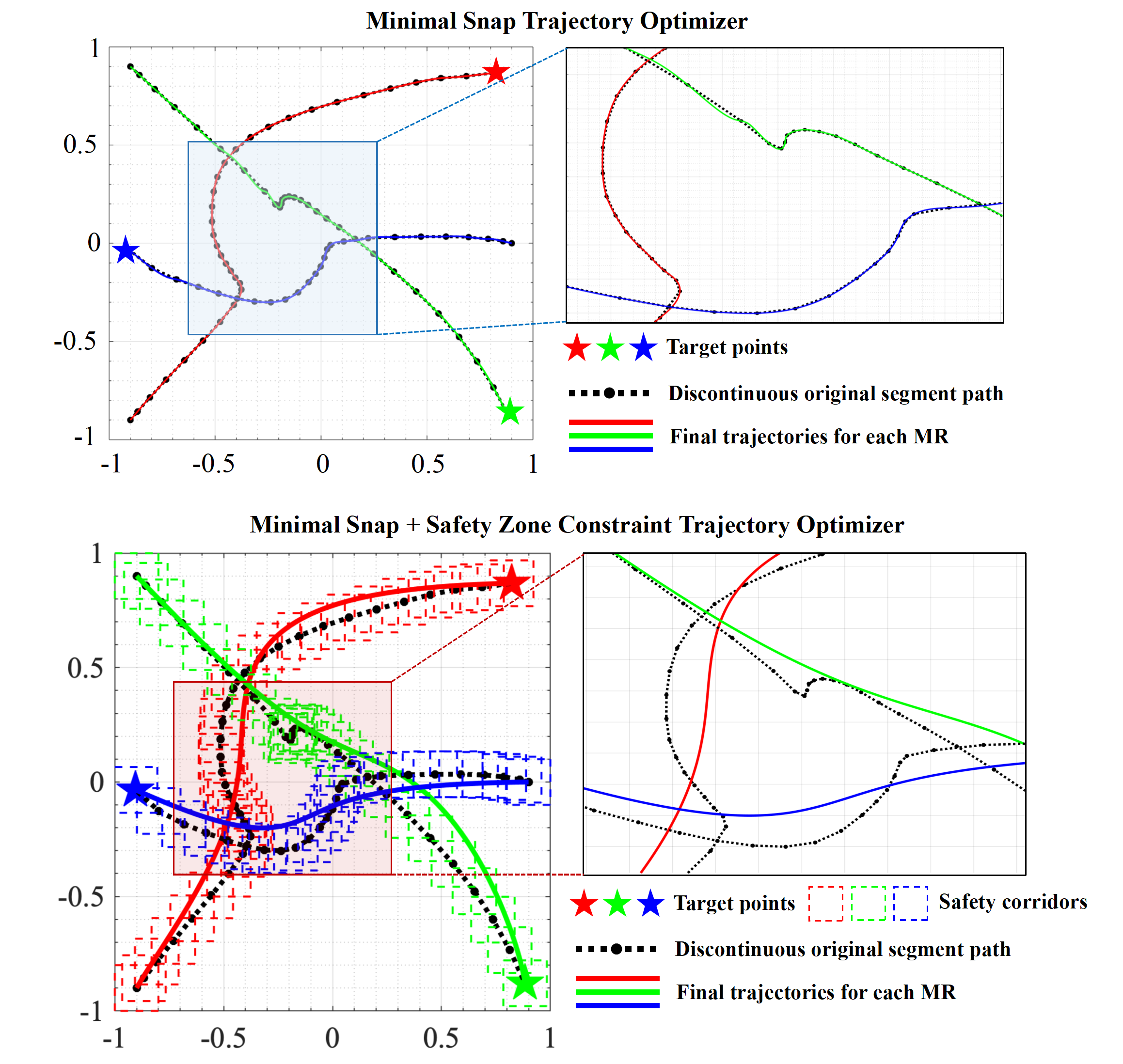}
	\caption{The final trajectory of cooperative motion planning process of multi-robots. The minimal snap-based trajectory optimizer at the back-end can produce a smoother robot trajectory and a more reasonable speed arrangement scheme.}
	\label{Fig_FinalTraj_Comparsion} 
\end{figure}  

\begin{figure}[!t]
	\centering
	\includegraphics[width=3.4 in]{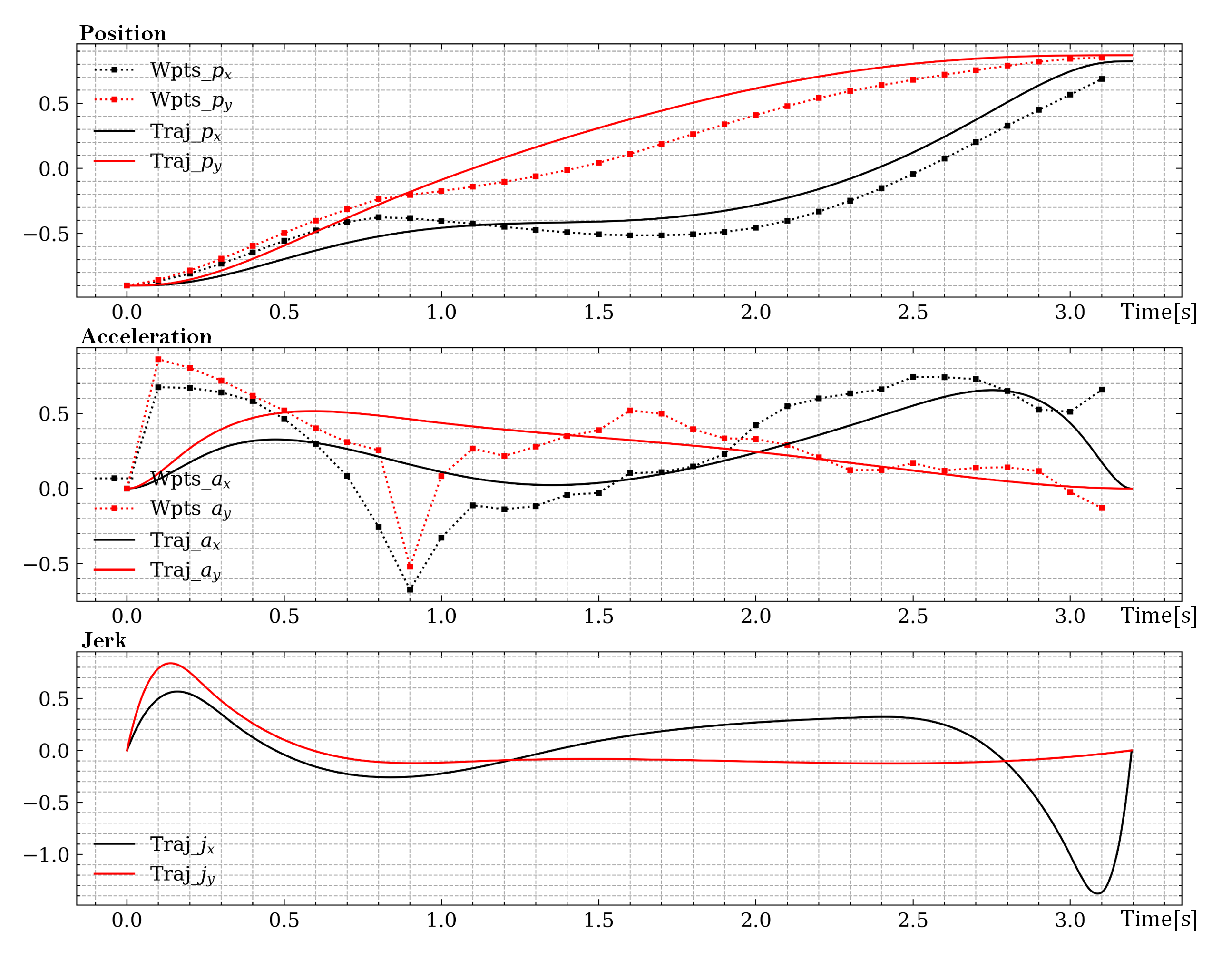}
	\caption{The visualization of multiple kino-dynamic profiles of one of the robots in the task scenario. The results suggest that after optimizing by our back-end module, the acceleration and jerk curves of the trajectory are smoother, more continuous,  and have better executability.}
	\label{Fig_AfterTrajOptim} 
\end{figure}  

%\begin{table*}[!t]
%	\centering
%	\caption{Performance metrics comparison of 3 robots collaborative path planning based on different algorithms}
%	\label{Tab_all_metrics}
%	\begin{tabular}{cccccc}
%		\toprule[1pt]
%		\textbf{Algorithms}                         & \multicolumn{5}{c}{\textbf{Different Performance Indicators}}                                                                         \\ \hline
%		\multicolumn{1}{l}{\textbf{Motion Planner}} & \textbf{Avg Total Distance} & \textbf{Avg Total Time} & \textbf{Avg Total Reward} & \textbf{Collsion Numbers} &\textbf{Success Rate} \\ \hline
%		ORCA                                        & 1.928                       & 114.538                 & -                         & 10(9049)                  & 73.627\%              \\
%		MATD3                                      & 3.497                       & 124.013                 & -28.804                   & 25(9673)                  & 92.301\%              \\
%		MAAC(CAS)                                   & 3.209                       & 123.679                 & -33.695                   & 14(9647)                  & 92.051\%              \\ \hline
%		\textbf{MASAC-auto}                         & \textbf{2.969}              & \textbf{118.897}        & \textbf{-39.545}          & \textbf{2(9274)}          & \textbf{93.592\%}     \\ \bottomrule[1pt]
%	\end{tabular}
%\end{table*}

It can be found that the final trajectory generated by the hybrid motion planner with safety zone constraints is better than that of the classical minimal snap method. Attributing to the implicit time allocation mechanism, the back-end trajectory optimization module helps the hybrid motion planner to generate a more reasonable speed arrangement scheme. This facilitates the correction of anomalous segments of the robot trajectory under the no-communication and local observation conditions. 

% 下图详细展现了加入后端优化模块前后位置、速度与加速度曲线的变化，虚线代表前段输出路径点的折线轨迹，实现为通过后端优化后的可执行轨迹。可以发现，最终轨迹速度与加速度曲线光滑且连续。并且，相较于速度可以发生突变的前端结果，最终轨迹的速度变化更加平缓，且速度与加速度都在机器人自身的限制范围之内。  
% 同时，我们的优化算法不仅保证了加速度输入在给定的限制的安全范围内，同时也能

Fig. \ref{Fig_AfterTrajOptim} presents more precisely the changes of the position profile,  the acceleration profile, and the jerk profile before and after introducing our back-end trajectory optimizer. The dashed lines represent the fold trajectory of the output of the front-end waypoints searching module, and the solid line is the final executable trajectory. It can be found that the position profile is fine-tuned by our back-end trajectory optimizer. Moreover, this hybrid planner can output smoother, continuous, and differentiable acceleration action curves than the original discontinuous and mutable acceleration curves. This performance ensures the stability and smoothness of robots during autonomous motion. Also, within the input safety range, the more stable acceleration change process helps to protect the actuator of the robot and thus save energy consumption. 
% Please add the following required packages to your document preamble:
% \usepackage[table,xcdraw]{xcolor}
% If you use beamer only pass "xcolor=table" option, i.e. \documentclass[xcolor=table]{beamer}
% Please add the following required packages to your document preamble:
% \usepackage[table,xcdraw]{xcolor}
% If you use beamer only pass "xcolor=table" option, i.e. \documentclass[xcolor=table]{beamer}

% Please add the following required packages to your document preamble:
% \usepackage{multirow}
\begin{table}[!t]
	\centering
	\caption{Performance comparison of different back-end trajectory generation methods. }
	\label{Table_Backend}
	\begin{tabular}{ccccc} 
		\toprule[1pt]
		\multirow{2}{*}{\textbf{Methods}} & \multicolumn{4}{c}{\textbf{Differenct Performance Indicators}}             \\ \cline{2-5} 
		& $\textbf{cost}_{\text{comfort}}$&$\textbf{cost}_{\text{smooth}}$&$\textbf{cost}_{\text{energy}}$&$\textbf{L}_{\text{distance}}$ \\ \hline
		Original Wpts                     & 1.053e-2          & -28.955           & 5.489            & 2.081            \\
		Cubic Spline                      & 1.886e-5          & -51.558           & 14.298           & 2.945            \\
		Minimal Snap                      & 1.501e-5          & \textbf{-109.914}          & 0.278            & 2.945            \\ \hline		\textbf{Ours}                              & \textbf{1.201e-6}          & -107.171          & \textbf{0.018}            & \textbf{2.824}            \\
		\bottomrule[1pt]		
	\end{tabular}
\end{table} 

Furthermore, we introduce four performance metrics to quantify the final performance of our method and conduct multiple experiments. The average results are summarized in Table \ref{Table_Backend}. The indicator for evaluating the  straightness of the final trajectory is as follows:
\begin{equation}
	\textbf{cost}_{\text{comfort}} = \sum_{i=1}^{n-1}(x_{i-1}+x_{i+1}-2x_i)^2+(y_{i-1}+y_{i+1}-2y_i)^2
\end{equation}
where $[x_{i-1},x_{i},x_{i+1}]$ and $[y_{i-1},y_i,y_{i+1}]$ are the horizontal and vertical coordinates of any three adjacent points $[P_{i-1},P_{i},P_{i+1}]$ in the final trajectory. $n$ represents the number of trajectory points after discretization. Also, the indicator for evaluating the smoothness of the final trajectory can be represented as follows:
\begin{equation}
	\begin{aligned}
	&\textbf{cost}_{\text{smoothness}}=  \\
	&-\sum_{i=1}^{n-1} \frac{\left(x_{i}-x_{i-1}\right) \left(x_{i+1}-x_{i}\right)+\left(y_{i}-y_{i-1}\right) \left(y_{i+1}-y_{i}\right)}{\|P_iP_{i-1}\|_{2}{\|P_{i+1}P_{i}\|_{2}}}
	\end{aligned}
\end{equation} 
$\textbf{cost}_\text{energy}=\sum_{i=1}^{n-1}(\bigtriangleup \bm{a})^2$ represents the force change during the robot movement, which can be used to measure energy consumption level. $\textbf{L}_\text{distance}$ represents the distance length.  Meanwhile, we select the ``front-end $+$ cubic spline" method and the ``front-end $+$  minimal snap" method as the benchmarks for comparison. The final results show that the output trajectories of our hybrid motion planner have better performance among the several groups given in this paper. 

\section{Conclusion}
In this paper, we propose a model-free multi-robot hybrid motion planner based on the MASAC-based waypoints searching method and the minimal snap with safety zone constraints trajectory optimizer.  This planer can output smooth, continuous, and dynamic feasible cooperative trajectories under non-communication and local observable conditions. In the front-end of the planner, we utilize MASAC with autotune exploration temperature coefficients to train multiple robots offline to learn to search available joint discrete waypoints.  In the back-end of the planner,  we construct a minimal snap optimization objective and introduce dynamic and safety constraints to revise and improve known discrete waypoints. By solving a  quadratic programming problem, we can obtain the final collision-free multi-robot executable trajectories. We set multi-group multi-robot motion planning experiment scenarios and select several mainstream baselines and an ideal algorithm under perfect perception assumption as the comparison benchmarks. The final simulation results verify the superior performance of our method.  Among multiple performance metrics, our method is closer to the ideal algorithm among several baselines.  Moreover, the final results also show that the back-end optimizer can successfully improve the quality of the final cooperative planning trajectories. 

% if have a single appendix:
%\appendix[Proof of the Zonklar Equations]
% or
%\appendix  % for no appendix heading
% do not use \section anymore after \appendix, only \section*
% is possibly needed

% use appendices with more than one appendix
% then use \section to start each appendix
% you must declare a \section before using any
% \subsection or using \label (\appendices by itself
% starts a section numbered zero.)
%

%\appendices
%\section{Proof of the First Zonklar Equation}
%Appendix one text goes here.
%
%% you can choose not to have a title for an appendix
%% if you want by leaving the argument blank
%\section{}
%Appendix two text goes here.
%
%
%% use section* for acknowledgment
%\section*{Acknowledgment}
%The authors would like to thank...

% Can use something like this to put references on a page
% by themselves when using endfloat and the captionsoff option.
\ifCLASSOPTIONcaptionsoff
  \newpage
\fi

% trigger a \newpage just before the given reference
% number - used to balance the columns on the last page
% adjust value as needed - may need to be readjusted if
% the document is modified later
%\IEEEtriggeratref{8}
% The "triggered" command can be changed if desired:
%\IEEEtriggercmd{\enlargethispage{-5in}}

% references section

% can use a bibliography generated by BibTeX as a .bbl file
% BibTeX documentation can be easily obtained at:
% http://mirror.ctan.org/biblio/bibtex/contrib/doc/
% The IEEEtran BibTeX style support page is at:
% http://www.michaelshell.org/tex/ieeetran/bibtex/
%\bibliographystyle{IEEEtran}
% argument is your BibTeX string definitions and bibliography database(s)
%\bibliography{IEEEabrv,../bib/paper}
%
% <OR> manually copy in the resultant .bbl file
% set second argument of \begin to the number of references
% (used to reserve space for the reference number labels box)
%\begin{thebibliography}{1}
%
%\bibitem{IEEEhowto:kopka}
%H.~Kopka and P.~W. Daly, \emph{A Guide to \LaTeX}, 3rd~ed.\hskip 1em plus
%  0.5em minus 0.4em\relax Harlow, England: Addison-Wesley, 1999.
%
%\end{thebibliography} 

\bibliographystyle{IEEEtran}
\bibliography{References.bib}

% biography section
% 
% If you have an EPS/PDF photo (graphicx package needed) extra braces are
% needed around the contents of the optional argument to biography to prevent
% the LaTeX parser from getting confused when it sees the complicated
% \includegraphics command within an optional argument. (You could create
% your own custom macro containing the \includegraphics command to make things
% simpler here.)
%\begin{IEEEbiography}[{\includegraphics[width=1in,height=1.25in,clip,keepaspectratio]{mshell}}]{Michael Shell}
% or if you just want to reserve a space for a photo:

%\begin{IEEEbiography}{Michael Shell}
%Biography text here.
%\end{IEEEbiography}

% if you will not have a photo at all:
%\begin{IEEEbiographynophoto}{John Doe}
%Biography text here.
%\end{IEEEbiographynophoto}

% insert where needed to balance the two columns on the last page with
% biographies
%\newpage

%\begin{IEEEbiographynophoto}{Jane Doe}
%Biography text here.
%\end{IEEEbiographynophoto}

% You can push biographies down or up by placing
% a \vfill before or after them. The appropriate
% use of \vfill depends on what kind of text is
% on the last page and whether or not the columns
% are being equalized.

%\vfill

% Can be used to pull up biographies so that the bottom of the last one
% is flush with the other column.
%\enlargethispage{-5in}

% that's all folks
\end{document}